\documentclass[11pt]{article}

\usepackage[final]{acl}

\usepackage{times}
\usepackage{latexsym}

\usepackage[T1]{fontenc}

\usepackage[utf8]{inputenc}

\usepackage{microtype}

\usepackage{inconsolata}

\usepackage{graphicx}
\usepackage[dvipsnames]{xcolor}
\usepackage{multirow}  %
\usepackage{makecell}
\usepackage{rotating}
\usepackage{float}
\usepackage{xargs}
\usepackage{booktabs}
\usepackage{amsmath}
\usepackage[table]{xcolor}
\usepackage[belowskip=-0.2em,labelfont=bf,labelsep=endash]{caption} %

\usepackage{pgf} %
\definecolor{high}{HTML}{019106}
\definecolor{mid}{HTML}{F7E379}
\definecolor{low}{HTML}{ec462e}
\newcommand{\opacity}{60}            

\newcommand{\lagom}{L\textsuperscript{\hspace{-3pt}A}G\textsubscript{O}M$\boldsymbol{\cdot}$NLP}

\newcommandx{\tgrad}[4][1=0.0, 2=0.5, 3=1.0]{%
	\ifdim #4 pt > #2 pt%
	\pgfmathparse{max(min(100.0*(#4-#2)/(#3-#2),100.0),0)}%
	\xdef\PercentColor{\pgfmathresult}%
	\cellcolor{high!\PercentColor!mid!\opacity}#4%
	\else
	\pgfmathparse{max(min(100.0*(#2-#4)/(#2-#1),100.0),0)}%
	\xdef\PercentColor{\pgfmathresult}%
	\cellcolor{low!\PercentColor!mid!\opacity}#4%
	\fi
}

\title{Typologically Informed Parameter Aggregation}

\author{
Stef Accou${}^{\diamond \dagger}$ \and
Wessel Poelman${}^\diamond$ \\
${}^\diamond$\lagom, Department of Computer Science, KU Leuven \\
${}^\dagger$Department of Linguistics, KU Leuven \\
\texttt{firstname.lastname@kuleuven.be}
}

\begin{document}

\maketitle
\begin{abstract}
Massively multilingual language models enable cross-lingual generalization but underperform on low-resource and unseen languages. While adapter-based fine-tuning offers a parameter-efficient solution, training language-specific adapters at scale remains costly. We introduce \emph{Typologically Informed Parameter Aggregation} (\textsc{TIPA}), a training-free method that constructs proxy language adapters by aggregating existing ones, weighted by typological similarity. Integrated into the MAD-X framework, these proxies enable zero-shot cross-lingual transfer without additional training. We evaluate \textsc{TIPA} on five NLP tasks and over 230 languages. \textsc{TIPA} consistently outperforms or matches baselines such as English-only fine-tuning or selecting the typologically closest language adapter. We see the largest gains for languages lacking dedicated adapters. Our results demonstrate that typologically informed aggregation provides a viable alternative to language-specific modules without any training needed.\footnote{Our code is available at: \href{https://github.com/stefaccou/TIPA}{github.com/stefaccou/TIPA}.}
\end{abstract}

\section{Introduction}
Massively multilingual language models have substantially expanded the reach of natural language processing (NLP), enabling zero-shot transfer across hundreds of languages. However, their performance remains uneven: while high-resource languages benefit from strong representations, models for under-resourced languages consistently underperform. This imbalance limits the inclusivity of NLP technologies and restricts their applicability for the majority of the world’s languages \cite{joshiStateFateLinguistic2020}.

Despite rapid advances, the development of state-of-the-art methods for cross-lingual transfer has so far been restricted to a relatively narrow set of languages. This limits claims about their generalizability, as they lack evidence from a typologically diverse sample. Cross-lingual transfer is especially relevant in low-resource settings, where adapting existing models to new languages can benefit from knowledge transferred from better-resourced ones.
Parameter-efficient fine-tuning (PEFT) approaches, particularly adapters \cite{houlsby2019parameterefficient}, provide a practical solution to adapting existing models. Adapter modules are lightweight, reusable components that can be inserted into pre-trained transformer models. In multilingual settings, language adapters have proven effective by improving performance with minimal training required \cite[\emph{e.g.,}][]{pfeifferMADXAdapterBasedFramework2020a,ustun2022udapter,klimaszewski2025no}.

While adapters are more efficient than full fine-tuning, they still require separate training for each language. Extending coverage to a broad set of languages remains difficult, especially since annotated data or monolingual corpora are scarce for certain languages.
What if we could instead re-use existing adapters and construct new ones for a new language?
This would be more scalable, foregoing the need for language-specific training.

\paragraph{Research Question.} 
We investigate whether linguistic information can be used to guide cross-lingual transfer without training language-specific adapter modules. Specifically, we ask: 

\vspace{-0.2em}
\begin{quote}
    \emph{Can we use typological information to construct proxy language adapters via parameter aggregation?}
\end{quote}
\vspace{-0.2em}

Our contributions are: (1) We introduce a typologically weighted aggregation method for constructing proxy language adapters in a training-free setting. By recombining existing adapters according to typology-based language proximity, we obtain stand-in modules that can be used in the MAD-X framework \cite{pfeifferMADXAdapterBasedFramework2020a}. (2) We evaluate the method and a selection of strong baselines on a sample of 234 languages over five downstream tasks. Our results highlight the scalability of the introduced parameter aggregation method and show that training-free combination of modules can work when using typological similarity as a prior.

\section{Related Work}

\paragraph{Cross-Lingual Transfer and PEFT.}
Multilingual models can be adapted to new languages.
Two popular ways of doing this is through continued pretraining, where the full weights of the model are updated, or through PEFT, where a small number of (additional) weights are updated.
Both approaches have been used to enable cross-lingual transfer.
Adapters have been successful in this area since they reduce computational costs while often preserving or increasing performance \cite{houlsby2019parameterefficient,ustun2020udapter,ustun2022udapter}. The MAD-X framework \cite{pfeifferMADXAdapterBasedFramework2020a, pfeiffer2020adapterhub,poth2023adapters} is particularly effective for cross-lingual transfer, offering modularity and composability across languages and tasks. While MAD-X-based approaches demonstrate the promise of modularity, they assume access to target-specific adapters, which can be costly to train.

\paragraph{Aggregation and Fusion Strategies.}
To extend coverage without per-language training, aggregation methods exploit the modular nature of adapters. Representation-level techniques such as AdapterFusion \cite{pfeiffer2021adapterfusion}, EMEA \cite{wangEfficientTestTime2021}, and more recently FLARE \cite{borchertLanguageFusionParameterEfficient2025} guide transfer by combining outputs from multiple adapters, at the cost of additional training and higher inference overhead. Parameter-level methods instead merge adapter weights directly, including AdapterSoup \cite{chronopoulouAdapterSoupWeightAveraging2023}, task arithmetic \cite{ilharco2023editing}, and recent language arithmetic approaches \cite{chronopoulou2024language, klimaszewski2025no}. These are training-free and inference-efficient but typically limited to pairwise or small-scale settings, often relying on carefully selected adapter combinations.

\paragraph{Typologically Guided Transfer.}
Linguistic typology is the study of structural commonalities and differences between languages.
Typological annotations can be collected in databases, which is the main way this information is used in NLP \cite{littell2017uriel,baylor2023presenta,haynie2023grambanks,khan2025uriel}.
These databases provide a feature vector per language, which allows for calculating distances between languages \cite{ploeger2024what,ploeger2025principled}. 
Since adapters also operate on the level of languages, typological feature vectors are a logical candidate for informing cross-lingual transfer. This has been used in adapter generation \cite{ustun2020udapter,ansellMADGMultilingualAdapter2021,ustun2022udapter} and in ensemble methods such as entropy-minimized aggregation \cite{wangEfficientTestTime2021}, showing that typological priors can be effective, particularly for low-resource languages.

Existing research shows the promise of modularity, parameter aggregation, and typologically-informed transfer. To our knowledge, no prior work combines these strands to construct training-free, proxy language adapters. Our method introduces a typologically weighted aggregation algorithm that leverages existing adapters to approximate language-specific modules without retraining.

\section{Methodology}
\begin{figure*}
    \centering
    \includegraphics[width=\textwidth]{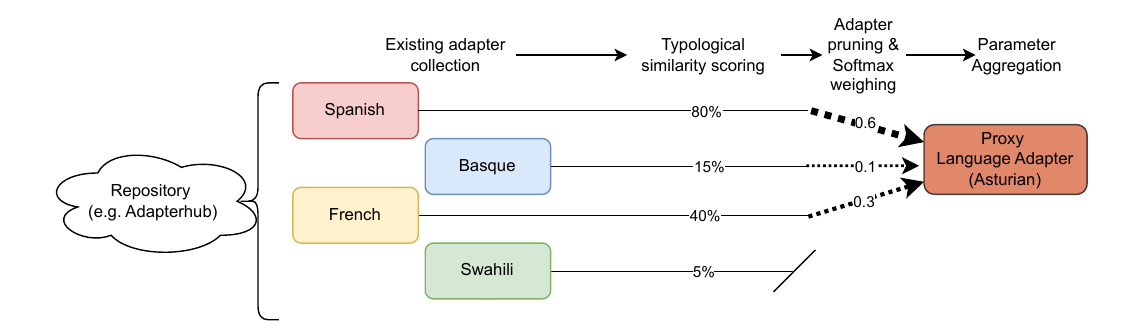}
    \caption{Typologically Informed Parameter Aggregation (\textsc{TIPA}): a training-free framework that constructs a  proxy language adapter for a target language by aggregating existing ones, weighted by typological similarity.}
    \label{fig:TIPA}
\end{figure*}

We build on the MAD-X framework \cite{pfeifferMADXAdapterBasedFramework2020a}. In this setting, a frozen multilingual transformer model is equipped with (1) a task adapter, trained on labeled data in a source language (typically English), and (2) a language adapter, trained on a language modeling task on a monolingual corpus. During training, the task adapter is stacked on top of the source language adapter. At inference time, the source language adapter can be substituted by a target language adapter, while keeping the task adapter fixed. This modularity enables separation of task and language knowledge.\footnote{Although it is disputed whether this actually works as well as reported \cite{kunz2024impact}.}

However, dedicated target-language adapters are not always available, especially for under-resourced languages. To address this, we introduce \emph{Typologically Informed Parameter Aggregation} (\textsc{TIPA}), a method for constructing stand-in language adapters by aggregating parameters from existing adapters. \textsc{TIPA} requires no additional training and can be applied to any language for which typological information is available. The resulting proxy adapter can then be integrated into the MAD-X architecture.

\paragraph{Adapter Resources.}
We assume access to a pool of pretrained language adapters, which serve as building blocks for the construction of the proxy adapter, and to task adapters trained on English task data. 
The only requirement is that the languages represented in the pool are also represented in the typological database used.

\paragraph{Typological Weighting.}
To approximate an adapter for target language $\ell_{\mathrm{tgt}}$, \textsc{TIPA} computes the proximity to the target for all source languages in the sample of available language adapters $\ell_{\mathrm{src}} \in \mathcal{L}$. Distances are derived from the URIEL+ database \cite{khan2025uriel}, which encodes a wide range of linguistic dimensions, including morphology, syntax, phonology, and character inventories. For each target language, pairwise distances ($d$) are normalized and converted into similarity weights through a softmax function:
\[
w_s = \frac{\exp(1 - d(\ell_{\mathrm{tgt}}, \ell_{\mathrm{src}}))}{\sum_{\ell_{\mathrm{src}}' \in \mathcal{L}} \exp(1 - d(\ell_{\mathrm{tgt}}, \ell_{\mathrm{src}}'))}
\]

\paragraph{Parameter Aggregation.}
Given the weighted set of source languages, \textsc{TIPA} constructs a new proxy adapter $L_{\mathrm{proxy}}$ through layer-wise parameter aggregation. Corresponding weight and bias matrices across all contributing language adapters are combined into a single module using the typologically deduced weights. The resulting proxy adapter matches the architecture of its source components and can thus be inserted directly into the MAD-X pipeline with the base model.  

\paragraph{Example.}
Table~\ref{tab:examples} shows some example proxy adapters for a given target language.
These are constructed using the adapters listed in Table~\ref{tab:existing-adapters}.
\begin{table}[h!]
    \centering
    \resizebox{\columnwidth}{!}{\begin{tabular}{l|lll}
    \toprule
     Target language & \multicolumn{3}{l}{Proxy created from (Top-3)}  \\
     \midrule
     Afrikaans & German (0.215) & Estonian (0.206) & Icelandic (0.195)\\
     Tibetan & Min Dong (0.204) & Javanese (0.204) &  Japanese (0.194) \\
     Catalan & Italian (0.215) & Spanish (0.199) & Greek (0.196) \\
     \bottomrule
\end{tabular}
}
    \caption{Top languages picked by \textsc{TIPA} to combine into a proxy adapter for the given target language.}
    \label{tab:examples}
\end{table}

\section{Evaluation}
We evaluate \textsc{TIPA} using XLM-RoBERTa base \cite{conneauUnsupervisedCrosslingualRepresentation2020} across five standard benchmarks: NER, POS tagging, COPA, QA, and topic classification, covering 234 languages. 
Detailed task descriptions, adapter resource configurations, and full language breakdowns are provided in \S\ref{sec:task_resource}.
We compare against the following baselines:

\begin{itemize}
    \item \textbf{English-only fine-tuning:} Fine-tune the model on English task data, then apply zero-shot to the target language. 
    \item \textbf{MAD-X with target-language adapter:} If available, a dedicated language adapter for the target language is used during inference. If no language adapter is available, we use the adapter trained on the typologically closest language.  
    \item \textbf{No Train but Gain:} A re-implementation of \cite{klimaszewski2025no}: additive combination of the English adapter with the typologically most similar available adapter. 
    \item \textbf{Uniform averaging:} Proxy adapters created by unweighted parameter averaging across all available language adapters in the sample.  

\end{itemize}

\section{Results}
\begin{table}[h!]
    \centering
    \resizebox{\columnwidth}{!}{\begin{tabular}{ll|lllll}
\toprule
 & ALL & NER & POS & \scalebox{0.8}{COPA} & QA & SIB \\
\cmidrule(lr){2-7}
\textit{\small \# Languages} &
\textit{\small (234)} &
\textit{\small (136)} &
\textit{\small (80)} &
\textit{\small (11)} &
\textit{\small (12)} &
\textit{\small (176)} \\
\midrule
Fine-tuning & \vtop{\hbox{\tgrad[43.1][45.4][54.1]{45.0}}\kern0.25ex\hbox{\scriptsize {\tiny$\pm$} 20.7}} & \vtop{\hbox{\tgrad[39.0][43.3][51.3]{39.0}}\kern0.25ex\hbox{\scriptsize {\tiny$\pm$} 18.4}} & \vtop{\hbox{\tgrad[38.9][45.4][46.9]{38.9}}\kern0.25ex\hbox{\scriptsize {\tiny$\pm$} 18.8}} & \textbf{\vtop{\hbox{\tgrad[50.3][51.8][55.5]{55.5}}\kern0.25ex\hbox{\scriptsize {\tiny$\pm$} 3.0}}} & \vtop{\hbox{\tgrad[53.4][72.1][72.9]{53.4}}\kern0.25ex\hbox{\scriptsize {\tiny$\pm$} 14.9}} & \vtop{\hbox{\tgrad[54.3][61.2][63.4]{61.2}}\kern0.25ex\hbox{\scriptsize {\tiny$\pm$} 25.0}} \\
MAD-X & \vtop{\hbox{\tgrad[43.1][45.4][54.1]{45.4}}\kern0.25ex\hbox{\scriptsize {\tiny$\pm$} 21.1}} & \vtop{\hbox{\tgrad[39.0][43.3][51.3]{43.3}}\kern0.25ex\hbox{\scriptsize {\tiny$\pm$} 21.4}} & \vtop{\hbox{\tgrad[38.9][45.4][46.9]{44.6}}\kern0.25ex\hbox{\scriptsize {\tiny$\pm$} 20.9}} & \vtop{\hbox{\tgrad[50.3][51.8][55.5]{52.0}}\kern0.25ex\hbox{\scriptsize {\tiny$\pm$} 2.1}} & \vtop{\hbox{\tgrad[53.4][72.1][72.9]{72.1}}\kern0.25ex\hbox{\scriptsize {\tiny$\pm$} 5.0}} & \vtop{\hbox{\tgrad[54.3][61.2][63.4]{56.5}}\kern0.25ex\hbox{\scriptsize {\tiny$\pm$} 24.3}} \\
No Train but Gain & \vtop{\hbox{\tgrad[43.1][45.4][54.1]{50.1}}\kern0.25ex\hbox{\scriptsize {\tiny$\pm$} 21.4}} & \vtop{\hbox{\tgrad[39.0][43.3][51.3]{49.3}}\kern0.25ex\hbox{\scriptsize {\tiny$\pm$} 18.1}} & \textbf{\vtop{\hbox{\tgrad[38.9][45.4][46.9]{46.9}}\kern0.25ex\hbox{\scriptsize {\tiny$\pm$} 20.5}}} & \vtop{\hbox{\tgrad[50.3][51.8][55.5]{50.3}}\kern0.25ex\hbox{\scriptsize {\tiny$\pm$} 1.4}} & \vtop{\hbox{\tgrad[53.4][72.1][72.9]{72.5}}\kern0.25ex\hbox{\scriptsize {\tiny$\pm$} 5.4}} & \vtop{\hbox{\tgrad[54.3][61.2][63.4]{61.6}}\kern0.25ex\hbox{\scriptsize {\tiny$\pm$} 25.1}} \\
Parameter averaging & \vtop{\hbox{\tgrad[43.1][45.4][54.1]{43.1}}\kern0.25ex\hbox{\scriptsize {\tiny$\pm$} 20.4}} & \vtop{\hbox{\tgrad[39.0][43.3][51.3]{39.5}}\kern0.25ex\hbox{\scriptsize {\tiny$\pm$} 18.1}} & \vtop{\hbox{\tgrad[38.9][45.4][46.9]{45.4}}\kern0.25ex\hbox{\scriptsize {\tiny$\pm$} 20.8}} & \vtop{\hbox{\tgrad[50.3][51.8][55.5]{51.5}}\kern0.25ex\hbox{\scriptsize {\tiny$\pm$} 2.0}} & \vtop{\hbox{\tgrad[53.4][72.1][72.9]{68.6}}\kern0.25ex\hbox{\scriptsize {\tiny$\pm$} 6.1}} & \vtop{\hbox{\tgrad[54.3][61.2][63.4]{54.3}}\kern0.25ex\hbox{\scriptsize {\tiny$\pm$} 25.1}} \\
TIPA & \textbf{\vtop{\hbox{\tgrad[43.1][45.4][54.1]{54.1}}\kern0.25ex\hbox{\scriptsize {\tiny$\pm$} 22.9}}} & \textbf{\vtop{\hbox{\tgrad[39.0][43.3][51.3]{51.3}}\kern0.25ex\hbox{\scriptsize {\tiny$\pm$} 18.5}}} & \vtop{\hbox{\tgrad[38.9][45.4][46.9]{46.8}}\kern0.25ex\hbox{\scriptsize {\tiny$\pm$} 20.7}} & \vtop{\hbox{\tgrad[50.3][51.8][55.5]{51.8}}\kern0.25ex\hbox{\scriptsize {\tiny$\pm$} 2.3}} & \textbf{\vtop{\hbox{\tgrad[53.4][72.1][72.9]{72.9}}\kern0.25ex\hbox{\scriptsize {\tiny$\pm$} 5.2}}} & \textbf{\vtop{\hbox{\tgrad[54.3][61.2][63.4]{63.4}}\kern0.25ex\hbox{\scriptsize {\tiny$\pm$} 24.3}}} \\
\bottomrule
\end{tabular}

}
    \caption[Overall results]{Baseline and \textsc{TIPA} scores aggregated across all tasks, alongside task-specific results. Scores are reported $\times$100. The color scheme situates relative performance per task. Full results for each task are in \S\ref{sec:distance_language_type}.}
    \label{tab:results_overview}
\end{table}

\begin{figure*}[t]
    \centering
    \includegraphics[width=\textwidth]{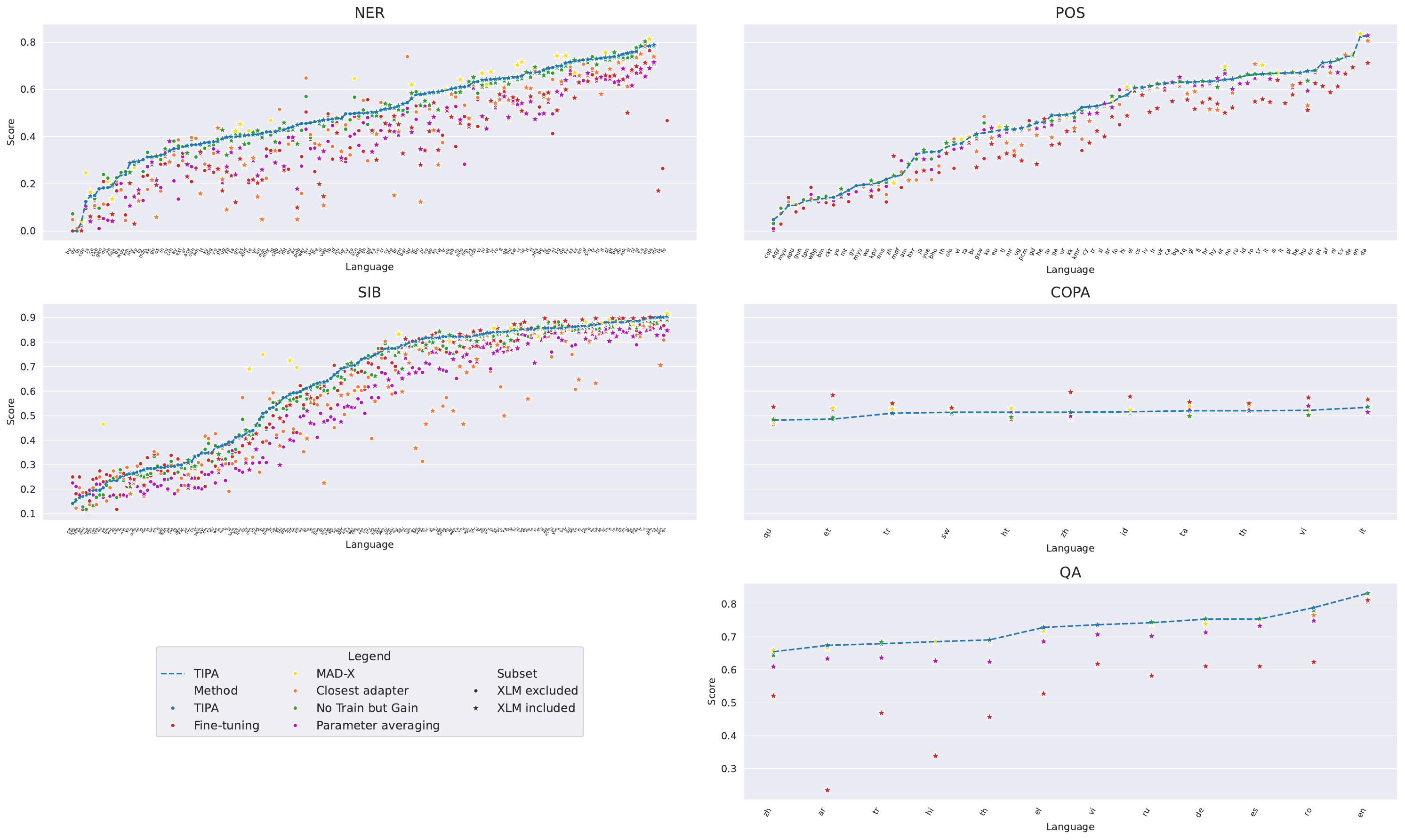}
    \caption{Scores for all tasks across available languages. We compare \textsc{TIPA} to \textit{No Train but Gain}, fine-tuning, uniform parameter averaging, and to either MAD-X or the typologically closest adapter, depending on adapter availability. Languages are marked for inclusion in XLM-R, and ordered based on increasing performance of our method. Larger formats of each presented figure can be found in \S\ref{sec:task_level_results}, alongside a more detailed analysis.}
    \label{fig:scatter}
\end{figure*}

\paragraph{Overall Results.}
As shown in Table~\ref{tab:results_overview}, \textsc{TIPA} outperforms all the evaluated baselines when aggregating results over all tasks.\footnote{Aggregating results like this has been rightfully criticized \cite{pikuliak2022average,ploeger2024what}, we do it here to show the viability of the method. In the following sections we provide more detailed analyses.}
The improvements are statistically significant under paired one–tailed t-tests (all comparisons $p<0.01$). Improvements are particularly pronounced for languages that lack a dedicated language adapter, where the proxy module directly fulfills its intended role as a stand-in within MAD-X. A more detailed breakdown by resource setting is presented in \S\ref{sec:task_level_results}. 
\textsc{TIPA} yields consistent gains relative to the widely used baseline of English-only fine-tuning, indicating that typology-aware parameter aggregation provides a useful prior compared to relying solely on the generalization capacity of the underlying model. \textsc{TIPA} improves over unweighted parameter averaging by +6.7\% on average (task-aggregated; $p<0.01$), confirming the value of linguistically informed weighting. Importantly, these improvements are achieved without any additional language-specific fine-tuning or adapter training.

\paragraph{Task-level Results.}
As shown in Figure~\ref{fig:scatter}, the benefits of \textsc{TIPA} are largest on morphologically sensitive tasks such as NER and POS. For higher-level semantic tasks (COPA, QA, SIB), \textsc{TIPA} remains competitive with the strongest adapter-based baselines and, in several settings, matches or surpasses the performance of dedicated pre-trained language adapters. 
Our re-implementation of \textit{No Train but Gain} \cite{klimaszewski2025no}, with automatic selection of the second adapter is consistently equalled or outperformed by \textsc{TIPA}, although it remains a strong second-best baseline. This confirms that, while other parameter aggregation methods can be competitive, aggregating information from multiple sources yields more robust gains than combining English with a single, typologically closest adapter.
Together, these results indicate that typologically-weighted aggregation is a viable, training-free alternative to per-language modules, with the strongest benefits in token-level tasks (Figure~\ref{fig:scatter}) and where dedicated adapters are unavailable. Nevertheless, performance of all methods varies greatly across tasks.
A more detailed overview is provided in \S\ref{sec:task_level_results}.

\paragraph{Feature Type Ablation and Pruning.}
The main parameters in \textsc{TIPA} specify which typological features are included in the distance calculation and how many adapters are considered in creating the proxy adapter.

We investigate which feature categories from URIEL+ are the most effective. We specifically consider: \emph{featural} (all available features), \emph{morphological}, and \emph{syntactic} for the language feature vectors. 
We find that for NER and POS, distances based on the \emph{morphological} category emerge as the most informative, while \emph{featural} remains the best overall choice.
The full results of this can be found in Table~\ref{tab:highest_score_overview}.

We also apply pruning strategies that either (1) retain the top-$k$ most similar source adapters for aggregation, instead of all available adapters or (2) only include adapters with a certain distance score to the target language based on a threshold.
We find there are some gains to be had from either pruning approach, but see no clear winner. 
Investigating more detailed refining of distance metrics and pruning methods is left for future work. Full results and further explanations are provided in \S\ref{sec:distance_language_type}.

\section{Conclusion}
We introduce \textsc{TIPA}, a training-free and parameter-efficient method for extending multilingual models to languages without dedicated adapters. By aggregating existing language adapters according to typological similarity, our method constructs proxy modules that integrate into the MAD-X framework. Evaluations across five tasks and more than 230 languages show improvements over strong baselines, with the largest gains for languages lacking adapters or excluded from pretraining.

Beyond gains on tasks, \textsc{TIPA} shows the benefit of using linguistic knowledge as an inexpensive but effective prior for cross-lingual transfer.

Typology-informed aggregation provides a practical way to improve cross-lingual transfer for underrepresented languages without increasing training or inference costs. In our experiments, this approach consistently benefited languages without model and adapter coverage. Future work should explore extending \textsc{TIPA} to multilingual models beyond XLM-R and investigating its use with task adapters and multi-task or full-parameter fine-tuning.

\newpage
\section*{Limitations}
Our findings have several limitations. 
First, performance varies substantially across tasks. 
While \textsc{TIPA} yields gains even for languages not included in pretraining, the performance of the underlying multilingual model deteriorates sharply for unseen scripts or unseen tokens, a limitation that our method cannot address due to the lack of meaningful representations.

Second, a key assumption of our approach is the availability of a set of language adapters for a given model. The effectiveness depends on the quality, breadth, and diversity of the pool of available adapters.
We did not explore these aspects in detail in the present study.

Third, although factors such as feature vector selection, similarity thresholds, or adapter selection can yield improvements, these factors were tuned heuristically. We did not perform an exhaustive search over these parameters.

Fourth, despite the inclusion of 234 languages (depending on the task), our study remains subject to bibliographic bias: sufficient corpora and evaluation datasets exist for only a fraction of the world’s languages \cite{rijkhoffLanguageSampling1998}. As a result, the reported gains for ``low-resource'' languages are only for those that have enough available resources for us to evaluate.

Finally, preliminary experiments on conditional multilingual language models (Gemma, Qwen) and PEFT methods (LoRa) did not see strong results. This suggests that the effectiveness of typology-aware aggregation may be architecture-dependent, and that further validation is needed across a broader range of model families.

\section*{Acknowledgments}
This paper builds on the results from the first author's Master Thesis, completed as part of the Master of Artificial Intellence programme at KU Leuven \cite{accouTIPATypologicallyInformed2025}. The authors want to thank Miryam de Lhoneux for her supervision and guidance throughout the research process. We also thank the anonymous reviewers for their constructive feedback.
WP is funded by a KU Leuven Bijzonder Onderzoeksfonds C1 project with reference C14/23/096.
The computational resources and services used were provided by the VSC (Flemish Supercomputer Center), funded by the Research Foundation - Flanders (FWO) and the Flemish Government - department EWI.

\vfill

\bibliography{custom}

\appendix

\newpage
\section{Appendix}
\label{sec:appendix}
\subsection{Tasks and Resource Settings}\label{sec:task_resource}

To ensure representativeness, we evaluate across five standard NLP benchmarks spanning multiple linguistic levels: 
\begin{itemize}
    \item NER \cite[WikiAnn;][]{rahimiMassivelyMultilingualTransfer2019}, a token-level classification task centered on identifying named entities.  
    \item POS tagging \cite[Universal Dependencies;][]{nivre2020universal}, a sequence-level annotation task assigning part-of-speech labels to words.  
    \item COPA \cite[XCOPA;][]{roemmeleChoicePlausibleAlternatives2011, pontiXCOPAMultilingualDataset2020}, a sentence-level reasoning task, requiring causal commonsense inference.  
    \item QA \cite[XQuAD;][]{rajpurkarSQuAD100000Questions2016, artetxeCrosslingualTransferabilityMonolingual2020}, a span-based question answering task.  
    \item Topic Classification \cite[SIB-200;][]{adelani2024sib200}, a simple document-level topic classification task across a broad sample of over 200 languages and dialects.  
\end{itemize}
Together these tasks cover token-level, sequence-level, and document-level evaluation. 
Our sample comprises 234 languages, including high-resource languages from XLM-R pre-training, languages with existing MAD-X adapters, and languages without dedicated adapters that thus require proxy reconstruction. NER, POS and SIB span the most typologically diverse samples of languages, serving as robust benchmarks for assessing the method's generalisability across languages with diverse resource scenarios. Table~\ref{tab:included_langs} summarises the number of languages per task and their distribution across these resource categories, and Tables~\ref{tab:included_langs_task} and \ref{tab:langs_single_tasks} give a full breakdown of languages included in each task.

\begin{table*}[ht]
	\centering
	\begin{tabular}{lc|cc|cc}
		\hline
		Task & Total languages & With adapter & Without adapter & In XLM-R & Not in XLM-R \\
		\hline
		NER  & 134 & 29 & 105 & 85 & 49 \\
		POS  &  80 & 19 &  61 & 57 & 23 \\
		COPA &  11 & 11 &   0 & 11 &  0 \\
		QA   &  12 & 11 &   1 & 12 &  0 \\
		SIB  & 176 & 25 & 151 & 81 & 95 \\
		\hline
		\textbf{Total} & \textbf{234} & \textbf{31} & \textbf{203} & \textbf{95} & \textbf{139} \\
		\hline
	\end{tabular}
	\caption[Evaluated languages per task]{Evaluated languages per task, along with resource-specific subsets.}
	\label{tab:included_langs}
\end{table*}

For each target language, we construct a proxy adapter using \textsc{TIPA} and combine it with the relevant task adapter under the MAD-X framework. Performance is measured on the test sets of each task. All models are evaluated without additional training, ensuring that results reflect the zero-shot generalisation capacity of proxy adapters.

\subsection{Task-level Results}\label{sec:task_level_results}
Figure~\ref{fig:scatter-one}compares \textsc{TIPA} to the baseline methods for all tasks and languages. The benefits of \textsc{TIPA} are largest on morphology-sensitive tasks such as NER and POS, where typological proximity provides informative guidance for reconstructing language-specific behaviour. In these cases, the improvements can be further augmented by limiting the similarity calculation process to only include morphological information. 
For NER, \textsc{TIPA} yields substantial gains, with particularly strong improvements for languages without a dedicated adapter, where it statistically outperforms all baselines ($p < 0.01$). In contrast, for languages with an existing adapter, performance is slightly below the MAD-X baseline. Further analysis shows that while higher token overlap consistently benefits all methods, \textsc{TIPA} achieves larger relative improvements in low-overlap settings. For POS, improvements are more modest and concentrated in languages natively supported by the underlying model.

For higher-level semantic tasks (COPA, QA, SIB), \textsc{TIPA} remains competitive with the strongest adapter-based baselines and, in most settings, matches or surpasses the performance of dedicated pre-trained language adapters. In COPA, variance across methods is smaller. Unlike in other tasks, fine-tuning achieves the strongest results for COPA, possibly due to the fact that all languages in the test set are natively supported high-resource languages.
For QA, \textsc{TIPA} statistically outperforms all methods ($p\leq0.05$), except for the No Train but Gain baseline ($p=0.10$), and even consistently outperforms the MAD-X baseline when a dedicated adapter is available. This indicates that for this reasoning task, updating the dedicated adapter with parameters from typologically related languages provides additional benefit over monolingual adapters.
Finally, for SIB, \textsc{TIPA} achieves the strongest overall results ($p<0.001$), significantly outperforming all baselines across conditions. The only exception is the subset of languages with a dedicated adapter, where performance falls short of the monolingual MAD-X adapter baseline.

\subsection{Distance Type and Language Pruning}\label{sec:distance_language_type}
All prior experiments employed the \textit{featural distance} metric from URIEL+ \cite{khan2025uriel} to calculate the typological similarity scores, which serve as the basis for weighting source language adapters in the approximation of target-language adapters. The featural distance metric is a composite of syntactic, morphological, phonological, and genetic information, offering the broadest coverage among available distance types. 
Also, it should be noted that in previous experiments, the full set of source adapters is weighted and integrated in the composite proxy adapter, regardless of their individual score. While more distant adapters receive a lower weight, they still contribute to the final proxy adapter.

To explore whether further gains in performance can be achieved within the \textsc{TIPA} method, we introduce two variations to the aggregation process:
\begin{enumerate}
	\item \textbf{Varying the distance type used for calculating adapter weights}\\ We evaluate if the use of morphological or syntactic distance has an impact on the score of the resulting reconstructed adapter. It should be noted that these distance types have narrower language coverage than the combined \textit{featural} metric. As a result, the number of evaluated languages is reduced from 234 to 207, with lower-resourced languages being disproportionately affected. The remaining distance types (such as phonological or genetic distance) provide even lower coverage, and are thus omitted from this analysis, as their inclusion would conflict with our goal of maintaining a maximally linguistically broad and inclusive evaluation setup.
	
	\item \textbf{Restricting the number of source adapters used in the reconstruction process}\\ We evaluate two strategies:
	\begin{enumerate}
		\item A fixed limit, where only the top five source languages (based on proximity for the different distance types) are retained during the softmax-weighted aggregation.
		\item A distance threshold, where only languages with a similarity score above 0.33 are considered and adequately weighted for inclusion in the reconstruction.
	\end{enumerate}
\end{enumerate}
\paragraph{Distance Types.}
For the NER and POS tagging tasks, morphological distance emerges as the most informative typological distance. In these tasks, it is the only distance type for which \textsc{TIPA} consistently outperforms all baselines with statistical significance ($p\leq0.05$), including the \textit{No Train but Gain} baseline, which could not be reliably surpassed under the default \textit{featural distance} setting. In contrast, syntactic distance proves less effective, yielding comparatively weaker results. For both methods, morphological distance just falls short of outperforming featural distance with statistical significance. Nevertheless, we consider morphological distance to be particularly relevant for tasks that rely more heavily on morphological cues, such as NER and POS.

For the COPA and QA tasks, the impact of distance type is minimal. Performance differences between morphological and syntactic distances remain within a margin of 0.2\%, rendering the variation negligible. Consequently, the results mirror those discussed in~\autoref{sec:task_level_results}, with no change in relative performance against the evaluated baselines.
For SIB, which is evaluated on a larger language sample, syntactic distance emerges as the most informative similarity criterium. It statistically outperforms all other distance types ($p\leq0.01$). Using syntactic distance in the approximation function results in a 0.5\% performance gain relative to featural distance on average. In contrast, morphological distance consistently underperforms. This can be due to the fact that morphological clues are more informative in surface-level tasks such as NER and POS, while syntactic similarity results in better transfer for higher-level tasks.

\paragraph{Limitation Strategies.}
Limitation strategies have a stronger impact on \textsc{TIPA} scores than distance type variations alone. Aggregating over all tasks, both the fixed-limit and the threshold-based strategies significantly outperform their corresponding unmodified variants ($p\leq0.03$). In particular, limiting the number of source adapters appears to be most effective when using the \textit{syntactic} distance metric. For all tasks, the best overall results achieved by our method are consistently associated with one of these limitation strategies, as shown in Table~\ref{tab:highest_score_overview}.

We leave the exploration of limit and threshold values for further research. The results presented here indicate that refining these parameters on a per-task basis may lead to further improvements. This suggests that our \textsc{TIPA} approach remains open to straightforward optimisations, and that even minor adjustments could yield additional gains beyond the current implementation in cross-lingual transfer settings.

\begin{table*}
	\centering
	\begin{minipage}{0.48\textwidth}
		\resizebox{!}{0.45\textheight}{\begin{tabular}{lrrrrr}
			\toprule
			Language & NER & COPA & POS & QA & SIB \\
			\midrule
			Achinese (ace) & \textcolor{ForestGreen}{1} & \textcolor{red}{0} & \textcolor{red}{0} & \textcolor{red}{0} & \textcolor{ForestGreen}{1} \\
			Afrikaans (af) & \textcolor{ForestGreen}{1} & \textcolor{red}{0} & \textcolor{ForestGreen}{1} & \textcolor{red}{0} & \textcolor{ForestGreen}{1} \\
			Albanian (sq) & \textcolor{ForestGreen}{1} & \textcolor{red}{0} & \textcolor{ForestGreen}{1} & \textcolor{red}{0} & \textcolor{red}{0} \\
			Amharic (am) & \textcolor{ForestGreen}{1} & \textcolor{red}{0} & \textcolor{ForestGreen}{1} & \textcolor{red}{0} & \textcolor{ForestGreen}{1} \\
			Arabic (ar) & \textcolor{ForestGreen}{1} & \textcolor{red}{0} & \textcolor{ForestGreen}{1} & \textcolor{ForestGreen}{1} & \textcolor{red}{0} \\
			Armenian (hy) & \textcolor{ForestGreen}{1} & \textcolor{red}{0} & \textcolor{ForestGreen}{1} & \textcolor{red}{0} & \textcolor{ForestGreen}{1} \\
			Assamese (as) & \textcolor{ForestGreen}{1} & \textcolor{red}{0} & \textcolor{red}{0} & \textcolor{red}{0} & \textcolor{ForestGreen}{1} \\
			Bambara (bm) & \textcolor{red}{0} & \textcolor{red}{0} & \textcolor{ForestGreen}{1} & \textcolor{red}{0} & \textcolor{ForestGreen}{1} \\
			Bashkir (ba) & \textcolor{ForestGreen}{1} & \textcolor{red}{0} & \textcolor{red}{0} & \textcolor{red}{0} & \textcolor{ForestGreen}{1} \\
			Basque (eu) & \textcolor{ForestGreen}{1} & \textcolor{red}{0} & \textcolor{ForestGreen}{1} & \textcolor{red}{0} & \textcolor{ForestGreen}{1} \\
			Belarusian (be) & \textcolor{ForestGreen}{1} & \textcolor{red}{0} & \textcolor{ForestGreen}{1} & \textcolor{red}{0} & \textcolor{ForestGreen}{1} \\
			Bengali (bn) & \textcolor{ForestGreen}{1} & \textcolor{red}{0} & \textcolor{red}{0} & \textcolor{red}{0} & \textcolor{ForestGreen}{1} \\
			Bhojpuri (bho) & \textcolor{red}{0} & \textcolor{red}{0} & \textcolor{ForestGreen}{1} & \textcolor{red}{0} & \textcolor{ForestGreen}{1} \\
			Bosnian (bs) & \textcolor{ForestGreen}{1} & \textcolor{red}{0} & \textcolor{red}{0} & \textcolor{red}{0} & \textcolor{ForestGreen}{1} \\
			Breton (br) & \textcolor{ForestGreen}{1} & \textcolor{red}{0} & \textcolor{ForestGreen}{1} & \textcolor{red}{0} & \textcolor{red}{0} \\
			Bulgarian (bg) & \textcolor{ForestGreen}{1} & \textcolor{red}{0} & \textcolor{ForestGreen}{1} & \textcolor{red}{0} & \textcolor{ForestGreen}{1} \\
			Catalan (ca) & \textcolor{ForestGreen}{1} & \textcolor{red}{0} & \textcolor{ForestGreen}{1} & \textcolor{red}{0} & \textcolor{ForestGreen}{1} \\
			Cebuano (ceb) & \textcolor{ForestGreen}{1} & \textcolor{red}{0} & \textcolor{red}{0} & \textcolor{red}{0} & \textcolor{ForestGreen}{1} \\
			Chinese (zh) & \textcolor{ForestGreen}{1} & \textcolor{ForestGreen}{1} & \textcolor{ForestGreen}{1} & \textcolor{ForestGreen}{1} & \textcolor{ForestGreen}{1} \\
			Crimean Tatar (crh) & \textcolor{ForestGreen}{1} & \textcolor{red}{0} & \textcolor{red}{0} & \textcolor{red}{0} & \textcolor{ForestGreen}{1} \\
			Croatian (hr) & \textcolor{ForestGreen}{1} & \textcolor{red}{0} & \textcolor{ForestGreen}{1} & \textcolor{red}{0} & \textcolor{ForestGreen}{1} \\
			Czech (cs) & \textcolor{ForestGreen}{1} & \textcolor{red}{0} & \textcolor{ForestGreen}{1} & \textcolor{red}{0} & \textcolor{ForestGreen}{1} \\
			Danish (da) & \textcolor{ForestGreen}{1} & \textcolor{red}{0} & \textcolor{ForestGreen}{1} & \textcolor{red}{0} & \textcolor{ForestGreen}{1} \\
			Dutch (nl) & \textcolor{ForestGreen}{1} & \textcolor{red}{0} & \textcolor{ForestGreen}{1} & \textcolor{red}{0} & \textcolor{ForestGreen}{1} \\
			Egyptian Arabic (arz) & \textcolor{ForestGreen}{1} & \textcolor{red}{0} & \textcolor{red}{0} & \textcolor{red}{0} & \textcolor{ForestGreen}{1} \\
			English (en) & \textcolor{ForestGreen}{1} & \textcolor{red}{0} & \textcolor{ForestGreen}{1} & \textcolor{ForestGreen}{1} & \textcolor{ForestGreen}{1} \\
			Esperanto (eo) & \textcolor{ForestGreen}{1} & \textcolor{red}{0} & \textcolor{red}{0} & \textcolor{red}{0} & \textcolor{ForestGreen}{1} \\
			Estonian (et) & \textcolor{ForestGreen}{1} & \textcolor{ForestGreen}{1} & \textcolor{ForestGreen}{1} & \textcolor{red}{0} & \textcolor{ForestGreen}{1} \\
			Faroese (fo) & \textcolor{ForestGreen}{1} & \textcolor{red}{0} & \textcolor{ForestGreen}{1} & \textcolor{red}{0} & \textcolor{ForestGreen}{1} \\
			Finnish (fi) & \textcolor{ForestGreen}{1} & \textcolor{red}{0} & \textcolor{ForestGreen}{1} & \textcolor{red}{0} & \textcolor{ForestGreen}{1} \\
			French (fr) & \textcolor{ForestGreen}{1} & \textcolor{red}{0} & \textcolor{ForestGreen}{1} & \textcolor{red}{0} & \textcolor{ForestGreen}{1} \\
			Friulian (fur) & \textcolor{ForestGreen}{1} & \textcolor{red}{0} & \textcolor{red}{0} & \textcolor{red}{0} & \textcolor{ForestGreen}{1} \\
			Galician (gl) & \textcolor{ForestGreen}{1} & \textcolor{red}{0} & \textcolor{ForestGreen}{1} & \textcolor{red}{0} & \textcolor{ForestGreen}{1} \\
			Georgian (ka) & \textcolor{ForestGreen}{1} & \textcolor{red}{0} & \textcolor{red}{0} & \textcolor{red}{0} & \textcolor{ForestGreen}{1} \\
			German (de) & \textcolor{ForestGreen}{1} & \textcolor{red}{0} & \textcolor{ForestGreen}{1} & \textcolor{ForestGreen}{1} & \textcolor{ForestGreen}{1} \\
			Greek (el) & \textcolor{ForestGreen}{1} & \textcolor{red}{0} & \textcolor{ForestGreen}{1} & \textcolor{ForestGreen}{1} & \textcolor{ForestGreen}{1} \\
			Guarani (gn) & \textcolor{ForestGreen}{1} & \textcolor{red}{0} & \textcolor{red}{0} & \textcolor{red}{0} & \textcolor{ForestGreen}{1} \\
			Gujarati (gu) & \textcolor{ForestGreen}{1} & \textcolor{red}{0} & \textcolor{red}{0} & \textcolor{red}{0} & \textcolor{ForestGreen}{1} \\
			Haitian Creole (ht) & \textcolor{red}{0} & \textcolor{ForestGreen}{1} & \textcolor{red}{0} & \textcolor{red}{0} & \textcolor{ForestGreen}{1} \\
			Hebrew (he) & \textcolor{ForestGreen}{1} & \textcolor{red}{0} & \textcolor{ForestGreen}{1} & \textcolor{red}{0} & \textcolor{ForestGreen}{1} \\
			Hindi (hi) & \textcolor{ForestGreen}{1} & \textcolor{red}{0} & \textcolor{ForestGreen}{1} & \textcolor{ForestGreen}{1} & \textcolor{ForestGreen}{1} \\
			Hungarian (hu) & \textcolor{ForestGreen}{1} & \textcolor{red}{0} & \textcolor{ForestGreen}{1} & \textcolor{red}{0} & \textcolor{ForestGreen}{1} \\
			Icelandic (is) & \textcolor{ForestGreen}{1} & \textcolor{red}{0} & \textcolor{ForestGreen}{1} & \textcolor{red}{0} & \textcolor{ForestGreen}{1} \\
			Igbo (ig) & \textcolor{ForestGreen}{1} & \textcolor{red}{0} & \textcolor{red}{0} & \textcolor{red}{0} & \textcolor{ForestGreen}{1} \\
			Ilocano (ilo) & \textcolor{ForestGreen}{1} & \textcolor{red}{0} & \textcolor{red}{0} & \textcolor{red}{0} & \textcolor{ForestGreen}{1} \\
			Indonesian (id) & \textcolor{ForestGreen}{1} & \textcolor{ForestGreen}{1} & \textcolor{ForestGreen}{1} & \textcolor{red}{0} & \textcolor{ForestGreen}{1} \\
			Irish (ga) & \textcolor{ForestGreen}{1} & \textcolor{red}{0} & \textcolor{ForestGreen}{1} & \textcolor{red}{0} & \textcolor{ForestGreen}{1} \\
			Italian (it) & \textcolor{ForestGreen}{1} & \textcolor{ForestGreen}{1} & \textcolor{ForestGreen}{1} & \textcolor{red}{0} & \textcolor{ForestGreen}{1} \\
			Japanese (ja) & \textcolor{ForestGreen}{1} & \textcolor{red}{0} & \textcolor{ForestGreen}{1} & \textcolor{red}{0} & \textcolor{ForestGreen}{1} \\
			Javanese (jv) & \textcolor{ForestGreen}{1} & \textcolor{red}{0} & \textcolor{red}{0} & \textcolor{red}{0} & \textcolor{ForestGreen}{1} \\
			Kannada (kn) & \textcolor{ForestGreen}{1} & \textcolor{red}{0} & \textcolor{red}{0} & \textcolor{red}{0} & \textcolor{ForestGreen}{1} \\
			Kazakh (kk) & \textcolor{ForestGreen}{1} & \textcolor{red}{0} & \textcolor{ForestGreen}{1} & \textcolor{red}{0} & \textcolor{ForestGreen}{1} \\
			Khmer (km) & \textcolor{ForestGreen}{1} & \textcolor{red}{0} & \textcolor{red}{0} & \textcolor{red}{0} & \textcolor{ForestGreen}{1} \\
			Kinyarwanda (rw) & \textcolor{ForestGreen}{1} & \textcolor{red}{0} & \textcolor{red}{0} & \textcolor{red}{0} & \textcolor{ForestGreen}{1} \\
			\bottomrule
		\end{tabular}
}
	\end{minipage}\hfill
	\begin{minipage}{0.48\textwidth}
		\resizebox{!}{0.45\textheight}{\begin{tabular}{lrrrrr}
			\toprule
			Language & NER & COPA & POS & QA & SIB \\
			\midrule
			Korean (ko) & \textcolor{ForestGreen}{1} & \textcolor{red}{0} & \textcolor{ForestGreen}{1} & \textcolor{red}{0} & \textcolor{ForestGreen}{1} \\
			Kyrgyz (ky) & \textcolor{ForestGreen}{1} & \textcolor{red}{0} & \textcolor{red}{0} & \textcolor{red}{0} & \textcolor{ForestGreen}{1} \\
			Latvian (lv) & \textcolor{ForestGreen}{1} & \textcolor{red}{0} & \textcolor{ForestGreen}{1} & \textcolor{red}{0} & \textcolor{red}{0} \\
			Lingala (ln) & \textcolor{ForestGreen}{1} & \textcolor{red}{0} & \textcolor{red}{0} & \textcolor{red}{0} & \textcolor{ForestGreen}{1} \\
			Lithuanian (lt) & \textcolor{ForestGreen}{1} & \textcolor{red}{0} & \textcolor{ForestGreen}{1} & \textcolor{red}{0} & \textcolor{ForestGreen}{1} \\
			Lombard (lmo) & \textcolor{ForestGreen}{1} & \textcolor{red}{0} & \textcolor{red}{0} & \textcolor{red}{0} & \textcolor{ForestGreen}{1} \\
			Luxembourgish (lb) & \textcolor{ForestGreen}{1} & \textcolor{red}{0} & \textcolor{red}{0} & \textcolor{red}{0} & \textcolor{ForestGreen}{1} \\
			Macedonian (mk) & \textcolor{ForestGreen}{1} & \textcolor{red}{0} & \textcolor{red}{0} & \textcolor{red}{0} & \textcolor{ForestGreen}{1} \\
			Malayalam (ml) & \textcolor{ForestGreen}{1} & \textcolor{red}{0} & \textcolor{red}{0} & \textcolor{red}{0} & \textcolor{ForestGreen}{1} \\
			Maltese (mt) & \textcolor{ForestGreen}{1} & \textcolor{red}{0} & \textcolor{ForestGreen}{1} & \textcolor{red}{0} & \textcolor{ForestGreen}{1} \\
			Maori (mi) & \textcolor{ForestGreen}{1} & \textcolor{red}{0} & \textcolor{red}{0} & \textcolor{red}{0} & \textcolor{ForestGreen}{1} \\
			Marathi (mr) & \textcolor{ForestGreen}{1} & \textcolor{red}{0} & \textcolor{ForestGreen}{1} & \textcolor{red}{0} & \textcolor{ForestGreen}{1} \\
			Minangkabau (min) & \textcolor{ForestGreen}{1} & \textcolor{red}{0} & \textcolor{red}{0} & \textcolor{red}{0} & \textcolor{ForestGreen}{1} \\
			Myanmar (Burmese) (my) & \textcolor{ForestGreen}{1} & \textcolor{red}{0} & \textcolor{red}{0} & \textcolor{red}{0} & \textcolor{ForestGreen}{1} \\
			Northern Kurdish (kmr) & \textcolor{red}{0} & \textcolor{red}{0} & \textcolor{ForestGreen}{1} & \textcolor{red}{0} & \textcolor{ForestGreen}{1} \\
			Norwegian (no) & \textcolor{ForestGreen}{1} & \textcolor{red}{0} & \textcolor{ForestGreen}{1} & \textcolor{red}{0} & \textcolor{red}{0} \\
			Occitan (oc) & \textcolor{ForestGreen}{1} & \textcolor{red}{0} & \textcolor{red}{0} & \textcolor{red}{0} & \textcolor{ForestGreen}{1} \\
			Polish (pl) & \textcolor{ForestGreen}{1} & \textcolor{red}{0} & \textcolor{ForestGreen}{1} & \textcolor{red}{0} & \textcolor{ForestGreen}{1} \\
			Portuguese (pt) & \textcolor{ForestGreen}{1} & \textcolor{red}{0} & \textcolor{ForestGreen}{1} & \textcolor{red}{0} & \textcolor{ForestGreen}{1} \\
			Punjabi (pa) & \textcolor{ForestGreen}{1} & \textcolor{red}{0} & \textcolor{red}{0} & \textcolor{red}{0} & \textcolor{ForestGreen}{1} \\
			Quechua (qu) & \textcolor{ForestGreen}{1} & \textcolor{ForestGreen}{1} & \textcolor{red}{0} & \textcolor{red}{0} & \textcolor{red}{0} \\
			Romanian (ro) & \textcolor{ForestGreen}{1} & \textcolor{red}{0} & \textcolor{ForestGreen}{1} & \textcolor{ForestGreen}{1} & \textcolor{ForestGreen}{1} \\
			Russian (ru) & \textcolor{ForestGreen}{1} & \textcolor{red}{0} & \textcolor{ForestGreen}{1} & \textcolor{ForestGreen}{1} & \textcolor{ForestGreen}{1} \\
			Scots Gaelic (gd) & \textcolor{ForestGreen}{1} & \textcolor{red}{0} & \textcolor{ForestGreen}{1} & \textcolor{red}{0} & \textcolor{ForestGreen}{1} \\
			Serbian (sr) & \textcolor{ForestGreen}{1} & \textcolor{red}{0} & \textcolor{ForestGreen}{1} & \textcolor{red}{0} & \textcolor{ForestGreen}{1} \\
			Sicilian (scn) & \textcolor{ForestGreen}{1} & \textcolor{red}{0} & \textcolor{red}{0} & \textcolor{red}{0} & \textcolor{ForestGreen}{1} \\
			Sindhi (sd) & \textcolor{ForestGreen}{1} & \textcolor{red}{0} & \textcolor{red}{0} & \textcolor{red}{0} & \textcolor{ForestGreen}{1} \\
			Sinhala (si) & \textcolor{ForestGreen}{1} & \textcolor{red}{0} & \textcolor{red}{0} & \textcolor{red}{0} & \textcolor{ForestGreen}{1} \\
			Slovenian (sl) & \textcolor{ForestGreen}{1} & \textcolor{red}{0} & \textcolor{ForestGreen}{1} & \textcolor{red}{0} & \textcolor{ForestGreen}{1} \\
			Somali (so) & \textcolor{ForestGreen}{1} & \textcolor{red}{0} & \textcolor{red}{0} & \textcolor{red}{0} & \textcolor{ForestGreen}{1} \\
			Sorani Kurdish (ckb) & \textcolor{ForestGreen}{1} & \textcolor{red}{0} & \textcolor{red}{0} & \textcolor{red}{0} & \textcolor{ForestGreen}{1} \\
			Spanish (es) & \textcolor{ForestGreen}{1} & \textcolor{red}{0} & \textcolor{ForestGreen}{1} & \textcolor{ForestGreen}{1} & \textcolor{ForestGreen}{1} \\
			Sundanese (su) & \textcolor{ForestGreen}{1} & \textcolor{red}{0} & \textcolor{red}{0} & \textcolor{red}{0} & \textcolor{ForestGreen}{1} \\
			Swahili (sw) & \textcolor{ForestGreen}{1} & \textcolor{ForestGreen}{1} & \textcolor{red}{0} & \textcolor{red}{0} & \textcolor{red}{0} \\
			Swedish (sv) & \textcolor{ForestGreen}{1} & \textcolor{red}{0} & \textcolor{ForestGreen}{1} & \textcolor{red}{0} & \textcolor{ForestGreen}{1} \\
			Tagalog (tl) & \textcolor{ForestGreen}{1} & \textcolor{red}{0} & \textcolor{ForestGreen}{1} & \textcolor{red}{0} & \textcolor{ForestGreen}{1} \\
			Tajik (tg) & \textcolor{ForestGreen}{1} & \textcolor{red}{0} & \textcolor{red}{0} & \textcolor{red}{0} & \textcolor{ForestGreen}{1} \\
			Tamil (ta) & \textcolor{ForestGreen}{1} & \textcolor{ForestGreen}{1} & \textcolor{ForestGreen}{1} & \textcolor{red}{0} & \textcolor{ForestGreen}{1} \\
			Tatar (tt) & \textcolor{ForestGreen}{1} & \textcolor{red}{0} & \textcolor{red}{0} & \textcolor{red}{0} & \textcolor{ForestGreen}{1} \\
			Telugu (te) & \textcolor{ForestGreen}{1} & \textcolor{red}{0} & \textcolor{ForestGreen}{1} & \textcolor{red}{0} & \textcolor{ForestGreen}{1} \\
			Thai (th) & \textcolor{ForestGreen}{1} & \textcolor{ForestGreen}{1} & \textcolor{ForestGreen}{1} & \textcolor{ForestGreen}{1} & \textcolor{ForestGreen}{1} \\
			Tibetan (bo) & \textcolor{ForestGreen}{1} & \textcolor{red}{0} & \textcolor{red}{0} & \textcolor{red}{0} & \textcolor{ForestGreen}{1} \\
			Tosk Albanian (als) & \textcolor{ForestGreen}{1} & \textcolor{red}{0} & \textcolor{red}{0} & \textcolor{red}{0} & \textcolor{ForestGreen}{1} \\
			Turkish (tr) & \textcolor{ForestGreen}{1} & \textcolor{ForestGreen}{1} & \textcolor{ForestGreen}{1} & \textcolor{ForestGreen}{1} & \textcolor{ForestGreen}{1} \\
			Turkmen (tk) & \textcolor{ForestGreen}{1} & \textcolor{red}{0} & \textcolor{red}{0} & \textcolor{red}{0} & \textcolor{ForestGreen}{1} \\
			Ukrainian (uk) & \textcolor{ForestGreen}{1} & \textcolor{red}{0} & \textcolor{ForestGreen}{1} & \textcolor{red}{0} & \textcolor{ForestGreen}{1} \\
			Urdu (ur) & \textcolor{ForestGreen}{1} & \textcolor{red}{0} & \textcolor{ForestGreen}{1} & \textcolor{red}{0} & \textcolor{ForestGreen}{1} \\
			Uyghur (ug) & \textcolor{ForestGreen}{1} & \textcolor{red}{0} & \textcolor{ForestGreen}{1} & \textcolor{red}{0} & \textcolor{ForestGreen}{1} \\
			Vietnamese (vi) & \textcolor{ForestGreen}{1} & \textcolor{ForestGreen}{1} & \textcolor{ForestGreen}{1} & \textcolor{ForestGreen}{1} & \textcolor{ForestGreen}{1} \\
			Waray (war) & \textcolor{ForestGreen}{1} & \textcolor{red}{0} & \textcolor{red}{0} & \textcolor{red}{0} & \textcolor{ForestGreen}{1} \\
			Welsh (cy) & \textcolor{ForestGreen}{1} & \textcolor{red}{0} & \textcolor{ForestGreen}{1} & \textcolor{red}{0} & \textcolor{ForestGreen}{1} \\
			Wolof (wo) & \textcolor{red}{0} & \textcolor{red}{0} & \textcolor{ForestGreen}{1} & \textcolor{red}{0} & \textcolor{ForestGreen}{1} \\
			Yoruba (yo) & \textcolor{ForestGreen}{1} & \textcolor{red}{0} & \textcolor{ForestGreen}{1} & \textcolor{red}{0} & \textcolor{ForestGreen}{1} \\
			Yue Chinese (yue) & \textcolor{red}{0} & \textcolor{red}{0} & \textcolor{ForestGreen}{1} & \textcolor{red}{0} & \textcolor{ForestGreen}{1} \\
			\bottomrule
		\end{tabular}
}
	\end{minipage}
	\caption{Language coverage per task.}
	\label{tab:included_langs_task}
\end{table*}
\begin{table*}
	\resizebox{\textwidth}{!}{\begin{tabular}{ll}
		\toprule
		Task & Languages \\
		\hline
		\midrule
		COPA & \makecell[l]{Akuntsu (aqz), Apurinã (apu), Chukot (ckt), Coptic (cop), Erzya (myv) \\ Komi-Zyrian (kpv), Livvi (olo), Manx (gv), Mbyá Guaraní (gun), Moksha (mdf) \\ Mundurukú (myu), Nigerian Pidgin (pcm), Russia Buriat (bxr), Skolt Sami (sms) \\ Swiss German (gsw), Tupinambá (tpn), Warlpiri (wbp)} \\
		\hline
		NER & \makecell[l]{Aragonese (an), Aymara (ay), Bavarian (bar), Chechen (ce), Chuvash (cv) \\ Corsican (co), Dhivehi (dv), Dimli (diq), Eastern Mari (mhr), Extremaduran (ext) \\ Frisian (fy), Gan Chinese (gan), Hakka Chinese (hak), Kurmanji Kurdish (ku) \\ Low German (nds), Mazanderani (mzn), Min Dong Chinese (cdo), Mingrelian (xmf) \\ Mongolian (mn), Neapolitan (nap), Nepali (ne), Northern Frisian (frr) \\ Ossetian (os), Pashto (ps), Romansh (rm), Scots (sco), Serbo-Croatian (sh) \\ Uzbek (uz), Veps (vep), Vlaams (vls), Western Panjabi (pnb), Wu Chinese (wuu) \\ Yakut (sah), Zeeuws (zea)} \\
		\hline
		POS & \makecell[l]{/} \\
		\hline
		QA & \makecell[l]{/} \\
		\hline
		SIB & \makecell[l]{Awadhi (awa), Ayacucho Quechua (quy), Balinese (ban), Bemba (bem) \\ Buginese (bug), Central Atlas Tamazight (tzm), Central Aymara (ayr) \\ Central Kanuri (knc), Chichewa (ny), Dari (prs), Dyula (dyu), Dzongkha (dz) \\ Eastern Yiddish (ydd), Ewe (ee), Fijian (fj), Fon (fon), Halh Mongolian (khk) \\ Hausa (ha), Iranian Persian (pes), Kabiyè (kbp), Kabuverdianu (kea) \\ Kabyle (kab), Kachin (kac), Kamba (kam), Kashmiri (ks), Kikuyu (ki), Lao (lo) \\ Luba-Lulua (lua), Luganda (lg), Luo (luo), Magahi (mag), Maithili (mai) \\ Meiteilon (Manipuri) (mni), Mesopotamian Arabic (acm), Mizo (lus) \\ Moroccan Arabic (ary), Mossi (mos), Najdi Arabic (ars), Nepali (npi) \\ Nigerian Fulfulde (fuv), North Azerbaijani (azj), North Levantine Arabic (apc) \\ Northern Uzbek (uzn), Norwegian Bokmål (nb), Nuer (nus), Odia (ory) \\ Pangasinan (pag), Papiamento (pap), Plateau Malagasy (plt), Rundi (rn) \\ Samoan (sm), Sango (sg), Santali (sat), Sepedi (nso), Sesotho (st), Shan (shn) \\ Shona (sn), South Azerbaijani (azb), Southern Pashto (pbt) \\ Standard Arabic (arb), Standard Malay (zsm), Swahili (swh), Swati (ss) \\ Tamasheq (taq), Tigrinya (ti), Tok Pisin (tpi), Tsonga (ts), Tswana (tn) \\ Tunisian Arabic (aeb), Twi (ak), Twi (tw), Umbundu (umb) \\ West Central Oromo (gaz), Xhosa (xh), Zulu (zu)} \\
		\hline
		\bottomrule
	\end{tabular}
}
	\caption{Languages included in only one task.}
	\label{tab:langs_single_tasks}
\end{table*}

\begin{table*}
    \centering
	\begin{tabular}{p{0.5\textwidth}}
\toprule
\multicolumn{1}{c}{\textbf{Languages with existing adapters}} \\
\midrule
Arabic, Basque, Chinese, Eastern Mari, English, Estonian, German, Greek, Guarani, 
Haitian Creole, Hindi, Icelandic, Ilocano, Indonesian, Italian, Japanese, Javanese, Maori, Min Dong Chinese,
Mingrelian, Myanmar (Burmese), Quechua, Russian, Serbian, Spanish, Swahili, Tamil, Thai, Turkish, Turkmen, Vietnamese \\
\bottomrule
\end{tabular}

	\caption[languages with existing XLM-R adapters]{Languages in our pool of existing MAD-X adapters trained for XLM-R. Retrieved from \url{https://huggingface.co/collections/AdapterHub/mad-x-adapters}}
	\label{tab:existing-adapters}
\end{table*}

\begin{figure*}
    \centering
    \includegraphics[width=\linewidth]{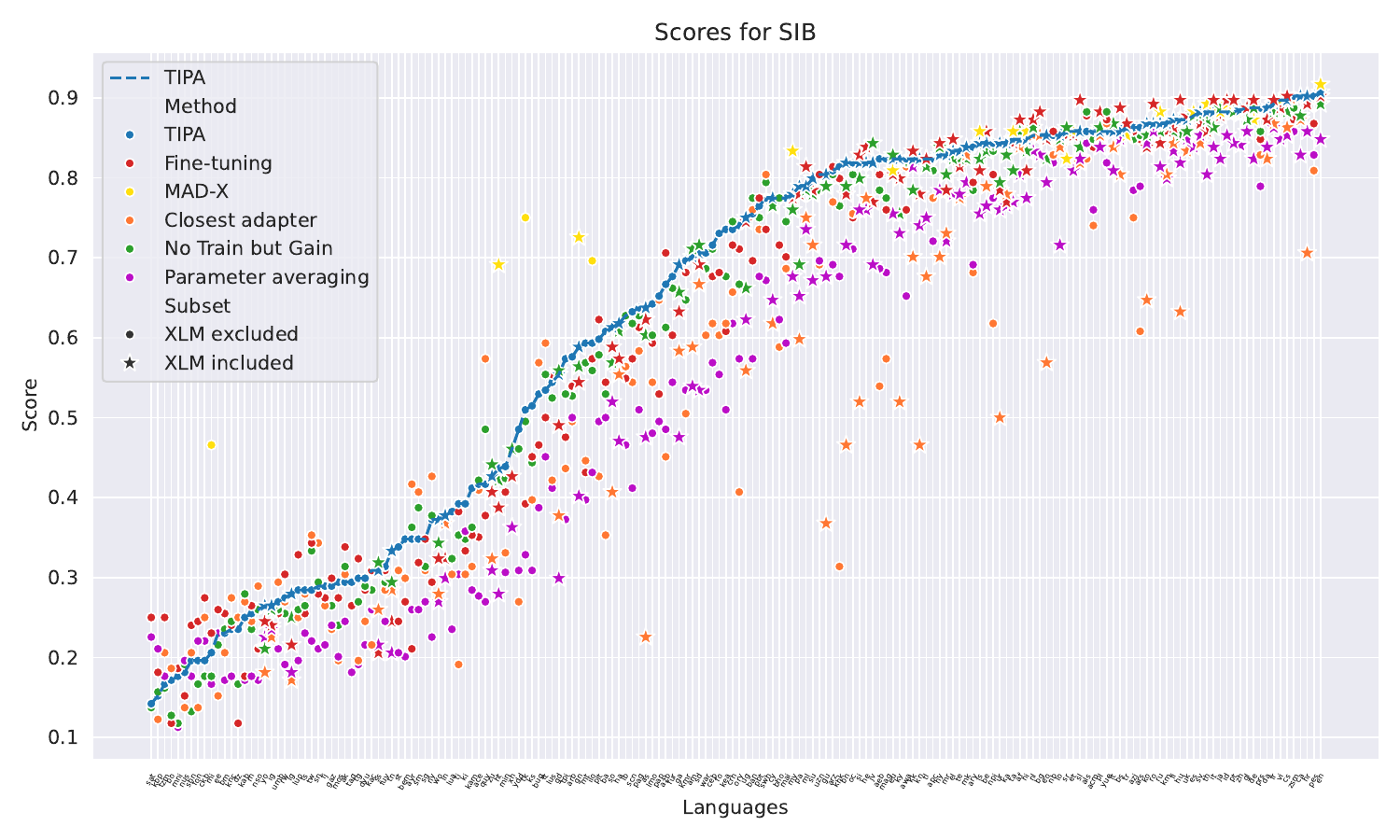}
    \includegraphics[width=\linewidth]{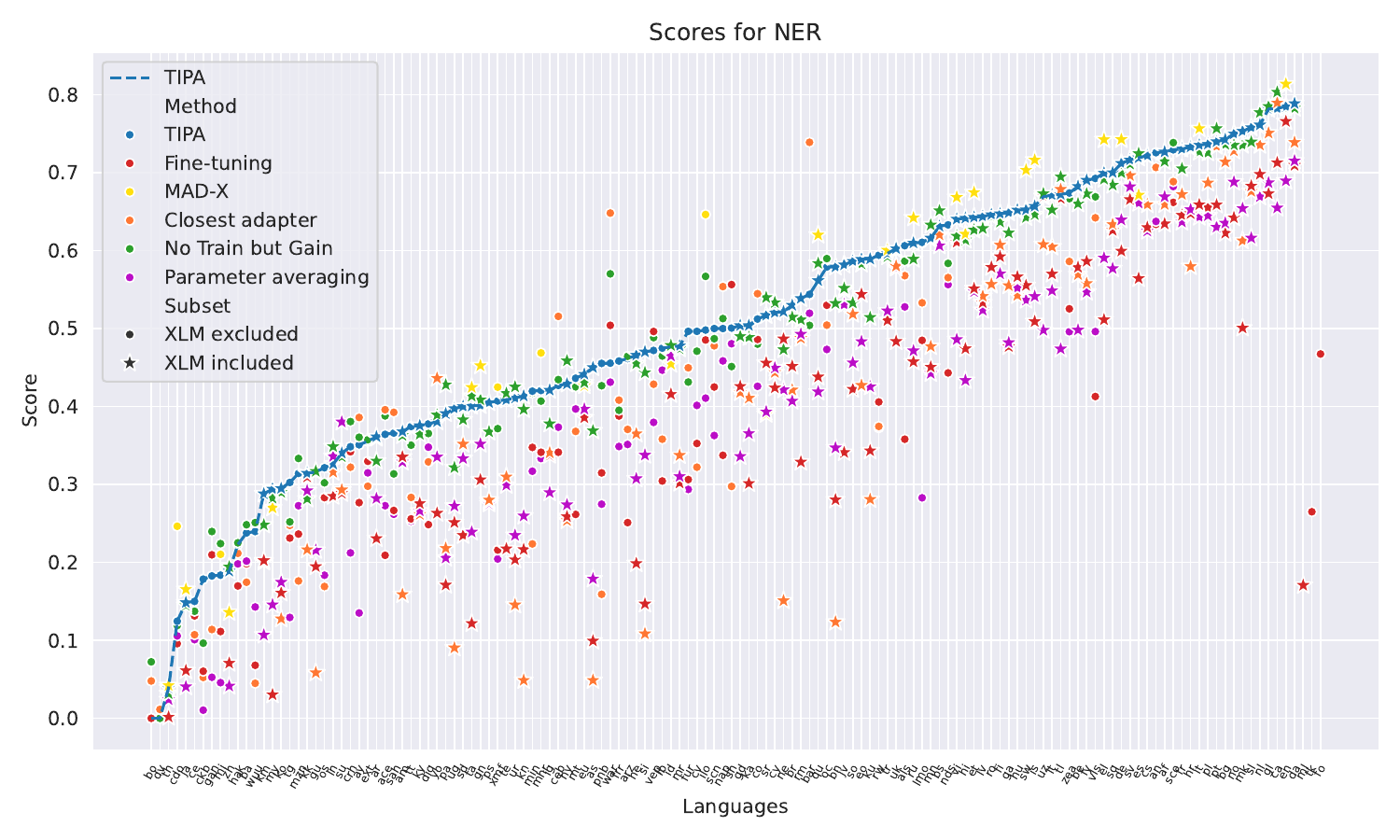}
    \caption{Scores for each task, across all evaluated languages. Comparing \textsc{TIPA} to the \textit{No Train but Gain} and fine-tuning baselines, as well as to either the MAD-X configuration or the typologically closest adapter, depending on adapter availability. Languages are presented in order of increasing performance of our method, and marked for native support in XLM-R. The results show that languages not included in XLM-R pre-training generally underperform, and that the benefits of our method primarily concern natively supported languages.}
    \label{fig:scatter-one}
\end{figure*}

\begin{figure*}\ContinuedFloat
    \centering
    \includegraphics[height=0.3\textheight]{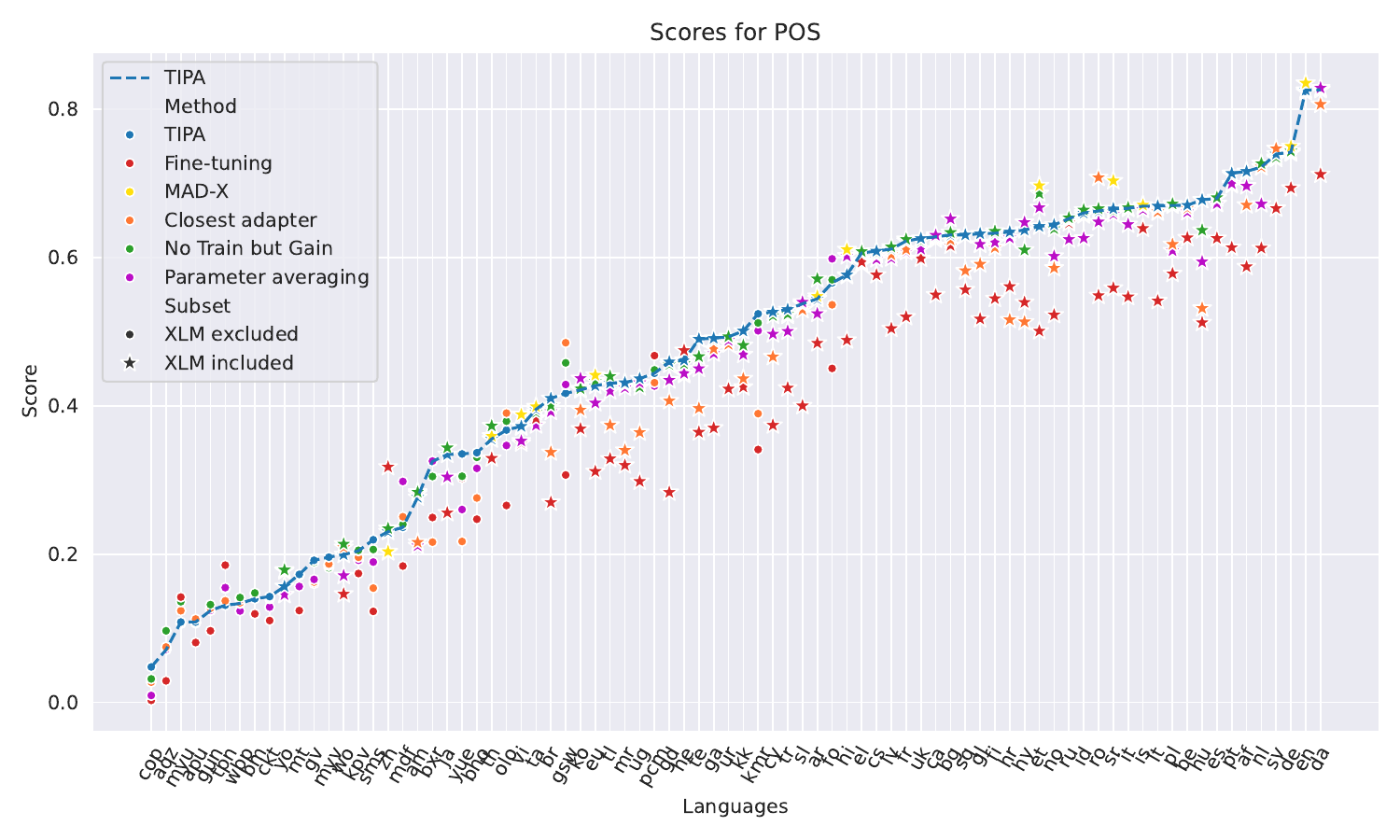}
    \includegraphics[height=0.299\textheight]{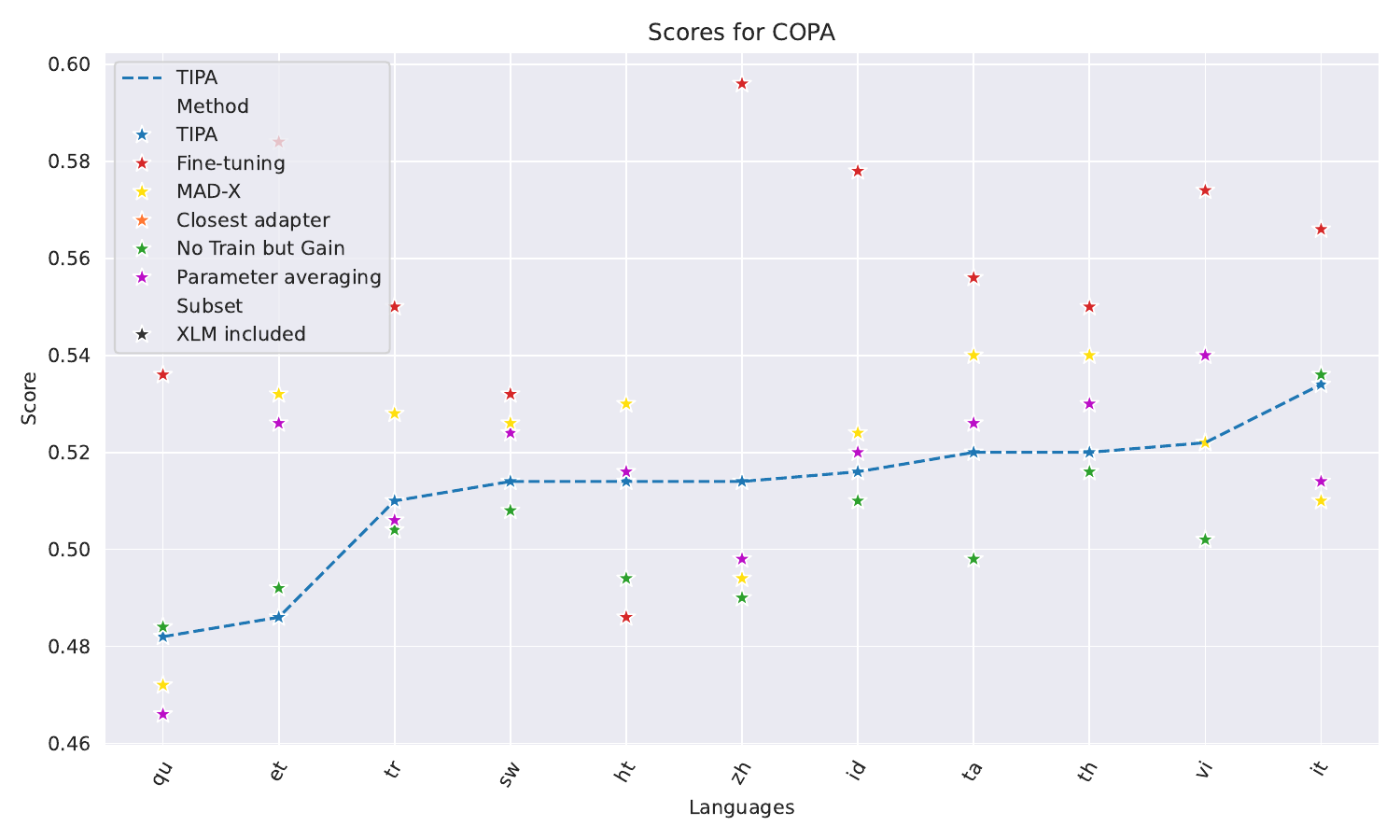}
    \includegraphics[height=0.298\textheight]{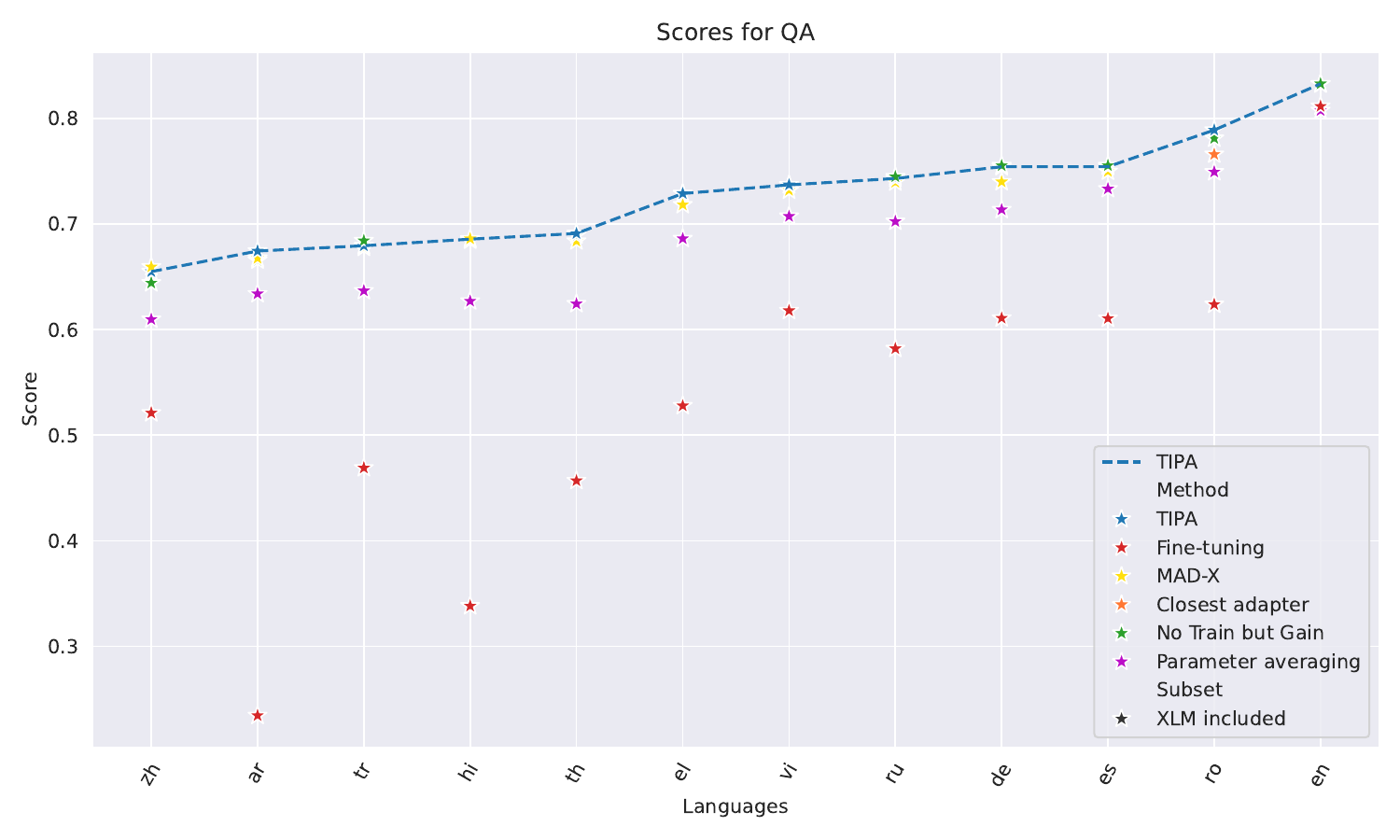}
    \caption{Scores for each task, across all evaluated languages. Comparing \textsc{TIPA} to the \textit{No Train but Gain} and fine-tuning baselines, as well as to either the MAD-X configuration or the typologically closest adapter, depending on adapter availability. Languages are presented in order of increasing performance of our method, and marked for native support in XLM-R. The results show that languages not included in XLM-R pre-training generally underperform, and that the benefits of our method primarily concern natively supported languages.}
    \label{fig:scatter-two}
\end{figure*}

\begin{table*}
	\begin{tabular}{lll|lllll}
\toprule
 &  & ALL & NER & POS & COPA & QA & SIB \\
SUBSET & METHOD &  &  &  &  &  &  \\
\midrule
\multirow[t]{7}{*}{Overall} & Fine-tuning & \tgrad[0.393][0.500][0.653]{0.450} & \tgrad[0.337][0.456][0.562]{0.390} & \tgrad[0.207][0.457][0.564]{0.389} & \textbf{\tgrad[0.503][0.519][0.555]{0.555}} & \tgrad[0.525][0.722][0.789]{0.534} & \tgrad[0.458][0.614][0.824]{0.612} \\
 & MAD-X & \tgrad[0.393][0.500][0.653]{0.454} & \tgrad[0.337][0.456][0.562]{0.433} & \tgrad[0.207][0.457][0.564]{0.446} & \tgrad[0.503][0.519][0.555]{0.520} & \tgrad[0.525][0.722][0.789]{0.721} & \tgrad[0.458][0.614][0.824]{0.565} \\
 & No Train but Gain & \tgrad[0.393][0.500][0.653]{0.501} & \tgrad[0.337][0.456][0.562]{0.493} & \textbf{\tgrad[0.207][0.457][0.564]{0.469}} & \tgrad[0.503][0.519][0.555]{0.503} & \tgrad[0.525][0.722][0.789]{0.725} & \tgrad[0.458][0.614][0.824]{0.616} \\
 & TIPA$_{Featural + limit}$ &  & \textbf{\tgrad[0.337][0.456][0.562]{0.513}} &  & \tgrad[0.503][0.519][0.555]{0.518} &  &  \\
 & TIPA$_{Featural}$ &  &  & \tgrad[0.207][0.457][0.564]{0.468} &  &  &  \\
 & TIPA$_{Morphological + threshold}$ &  &  &  &  & \textbf{\tgrad[0.525][0.722][0.789]{0.729}} &  \\
 & TIPA$_{Syntactic + limit}$ & \textbf{\tgrad[0.393][0.500][0.653]{0.541}} &  &  &  &  & \textbf{\tgrad[0.458][0.614][0.824]{0.634}} \\
\cline{1-8}
\multirow[t]{7}{*}{No adapter} & Fine-tuning & \tgrad[0.393][0.500][0.653]{0.447} & \tgrad[0.337][0.456][0.562]{0.395} & \tgrad[0.207][0.457][0.564]{0.367} &  & \tgrad[0.525][0.722][0.789]{0.624} & \tgrad[0.458][0.614][0.824]{0.585} \\
 & MAD-X & \tgrad[0.393][0.500][0.653]{0.431} & \tgrad[0.337][0.456][0.562]{0.414} & \tgrad[0.207][0.457][0.564]{0.410} &  & \tgrad[0.525][0.722][0.789]{0.766} & \tgrad[0.458][0.614][0.824]{0.522} \\
 & No Train but Gain & \tgrad[0.393][0.500][0.653]{0.491} & \tgrad[0.337][0.456][0.562]{0.496} & \tgrad[0.207][0.457][0.564]{0.439} &  & \tgrad[0.525][0.722][0.789]{0.781} & \tgrad[0.458][0.614][0.824]{0.590} \\
 & TIPA$_{Featural + limit}$ & \textbf{\tgrad[0.393][0.500][0.653]{0.520}} & \textbf{\tgrad[0.337][0.456][0.562]{0.516}} &  &  &  &  \\
 & TIPA$_{Featural}$ &  &  & \textbf{\tgrad[0.207][0.457][0.564]{0.440}} &  &  &  \\
 & TIPA$_{Morphological}$ &  &  &  &  & \textbf{\tgrad[0.525][0.722][0.789]{0.789}} &  \\
 & TIPA$_{Syntactic + limit}$ &  &  &  &  &  & \textbf{\tgrad[0.458][0.614][0.824]{0.608}} \\
\cline{1-8}
\multirow[t]{7}{*}{With adapter} & Fine-tuning & \tgrad[0.393][0.500][0.653]{0.500} & \tgrad[0.337][0.456][0.562]{0.371} & \tgrad[0.207][0.457][0.564]{0.507} & \textbf{\tgrad[0.503][0.519][0.555]{0.555}} & \tgrad[0.525][0.722][0.789]{0.525} & \tgrad[0.458][0.614][0.824]{0.767} \\
 & MAD-X & \tgrad[0.393][0.500][0.653]{0.605} & \tgrad[0.337][0.456][0.562]{0.505} & \tgrad[0.207][0.457][0.564]{0.561} & \tgrad[0.503][0.519][0.555]{0.520} & \tgrad[0.525][0.722][0.789]{0.717} & \textbf{\tgrad[0.458][0.614][0.824]{0.824}} \\
 & No Train but Gain & \tgrad[0.393][0.500][0.653]{0.568} & \tgrad[0.337][0.456][0.562]{0.484} & \tgrad[0.207][0.457][0.564]{0.562} & \tgrad[0.503][0.519][0.555]{0.503} & \tgrad[0.525][0.722][0.789]{0.720} & \tgrad[0.458][0.614][0.824]{0.773} \\
 & TIPA$_{Featural + limit}$ &  &  &  & \tgrad[0.503][0.519][0.555]{0.518} & \textbf{\tgrad[0.525][0.722][0.789]{0.723}} &  \\
 & TIPA$_{Morphological + limit}$ &  &  &  &  &  & \tgrad[0.458][0.614][0.824]{0.792} \\
 & TIPA$_{Morphological + threshold}$ &  &  & \textbf{\tgrad[0.207][0.457][0.564]{0.564}} &  &  &  \\
 & TIPA$_{Syntactic + limit}$ & \textbf{\tgrad[0.393][0.500][0.653]{0.621}} & \textbf{\tgrad[0.337][0.456][0.562]{0.509}} &  &  &  &  \\
\cline{1-8}
\multirow[t]{8}{*}{XLM included} & Fine-tuning & \tgrad[0.393][0.500][0.653]{0.530} & \tgrad[0.337][0.456][0.562]{0.425} & \tgrad[0.207][0.457][0.564]{0.485} & \textbf{\tgrad[0.503][0.519][0.555]{0.555}} & \tgrad[0.525][0.722][0.789]{0.534} & \tgrad[0.458][0.614][0.824]{0.754} \\
 & MAD-X & \tgrad[0.393][0.500][0.653]{0.542} & \tgrad[0.337][0.456][0.562]{0.469} & \tgrad[0.207][0.457][0.564]{0.537} & \tgrad[0.503][0.519][0.555]{0.520} & \tgrad[0.525][0.722][0.789]{0.721} & \tgrad[0.458][0.614][0.824]{0.691} \\
 & No Train but Gain & \tgrad[0.393][0.500][0.653]{0.605} & \tgrad[0.337][0.456][0.562]{0.542} & \tgrad[0.207][0.457][0.564]{0.560} & \tgrad[0.503][0.519][0.555]{0.503} & \tgrad[0.525][0.722][0.789]{0.725} & \tgrad[0.458][0.614][0.824]{0.754} \\
 & TIPA$_{Featural + limit}$ &  & \textbf{\tgrad[0.337][0.456][0.562]{0.562}} &  & \tgrad[0.503][0.519][0.555]{0.518} &  &  \\
 & TIPA$_{Featural + threshold}$ &  &  & \textbf{\tgrad[0.207][0.457][0.564]{0.562}} &  &  &  \\
 & TIPA$_{Featural}$ &  &  &  &  &  & \textbf{\tgrad[0.458][0.614][0.824]{0.768}} \\
 & TIPA$_{Morphological + threshold}$ &  &  &  &  & \textbf{\tgrad[0.525][0.722][0.789]{0.729}} &  \\
 & TIPA$_{Syntactic}$ & \textbf{\tgrad[0.393][0.500][0.653]{0.653}} &  &  &  &  &  \\
\cline{1-8}
\multirow[t]{6}{*}{XLM excluded} & Fine-tuning & \tgrad[0.393][0.500][0.653]{0.401} & \tgrad[0.337][0.456][0.562]{0.337} & \tgrad[0.207][0.457][0.564]{0.207} &  &  & \tgrad[0.458][0.614][0.824]{0.505} \\
 & MAD-X & \tgrad[0.393][0.500][0.653]{0.393} & \tgrad[0.337][0.456][0.562]{0.371} & \tgrad[0.207][0.457][0.564]{0.221} &  &  & \tgrad[0.458][0.614][0.824]{0.458} \\
 & No Train but Gain & \tgrad[0.393][0.500][0.653]{0.429} & \tgrad[0.337][0.456][0.562]{0.407} & \tgrad[0.207][0.457][0.564]{0.243} &  &  & \tgrad[0.458][0.614][0.824]{0.498} \\
 & TIPA$_{Morphological + threshold}$ &  &  & \textbf{\tgrad[0.207][0.457][0.564]{0.255}} &  &  &  \\
 & TIPA$_{Syntactic + limit}$ & \textbf{\tgrad[0.393][0.500][0.653]{0.468}} &  &  &  &  & \textbf{\tgrad[0.458][0.614][0.824]{0.534}} \\
 & TIPA$_{Syntactic + threshold}$ &  & \textbf{\tgrad[0.337][0.456][0.562]{0.443}} &  &  &  &  \\
\cline{1-8}
\bottomrule
\end{tabular}

	\caption[Best performing \textsc{TIPA} variations]{Baseline scores and best scores obtained for each subset and task. The highest score for each task within each subset of languages is put in bold, and the colour scheme situates the relative performance of the scores per task. The highest scoring method from our framework is presented, marked for distance type and limiting method, along with its score in the relevant task. The \textit{ALL} column presents task-aggregated results, in which task scores are weighted based on the number of evaluated languages.}
	\label{tab:highest_score_overview}
\end{table*}

\end{document}